\documentclass[letterpaper]{article} 
\usepackage{aaai25}  
\usepackage{times}  
\usepackage{helvet}  
\usepackage{courier}  
\usepackage[hyphens]{url}  
\usepackage{graphicx} 
\usepackage{natbib}  
\usepackage{caption} 
\usepackage{algorithm}
\usepackage{algorithmic}

\usepackage[table]{xcolor}

\usepackage{times}  
\usepackage{helvet}  
\usepackage{courier}  
\usepackage[hyphens]{url}  
\usepackage{graphicx} 
\usepackage{natbib}  
\usepackage{caption} 
\usepackage{algorithm}
\usepackage{algorithmic}

\usepackage{amsfonts}
\usepackage{latexsym}
\usepackage{amssymb}
\usepackage{amsmath}
\usepackage{amsthm}
\usepackage{booktabs}
\usepackage{enumitem}
\usepackage{graphicx}
\usepackage{color}
\usepackage{overpic}
\usepackage{multirow}
\usepackage{xcolor}         
\usepackage{colortbl}
\usepackage{bbding}
\usepackage{pifont}
\usepackage{fontawesome}
\usepackage{subcaption}

\definecolor{lightgrey}{RGB}{244,244,244}
\definecolor{grey}{RGB}{128,128,128}
\definecolor{midgrey}{RGB}{225,225,225}
\definecolor{forestgreen}{RGB}{47, 159, 87}
\newcommand{\cmark}{\color{forestgreen}\ding{51}}%
\newcommand{\xmark}{\color{red}\ding{55}}%

\usepackage[capitalize]{cleveref}
\crefname{section}{Sec.}{Secs.}
\Crefname{section}{Section}{Sections}
\Crefname{table}{Table}{Tables}

\usepackage{soul}
\graphicspath{{figures/}}

\definecolor{Gray}{gray}{0.95}
\definecolor{Cyan}{rgb}{0.88,1,1}

\usepackage{newfloat}
\usepackage{listings}
\DeclareCaptionStyle{ruled}{labelfont=normalfont,labelsep=colon,strut=off} 
\floatstyle{ruled}
\newfloat{listing}{tb}{lst}{}
\floatname{listing}{Listing}
\nocopyright

\title{Envisioning Class Entity Reasoning by Large Language \\ Models for Few-shot Learning}
\author {
    Mushui Liu, Fangtai Wu, Bozheng Li, Ziqian Lu, Yunlong Yu\thanks{Corresponding Author}, Xi Li
}
\affiliations {
     Zhejiang University
}

\usepackage{bibentry}

\begin{document}

\maketitle
\begin{abstract}
Few-shot learning (FSL) aims to recognize new concepts using a limited number of visual samples. Existing approaches attempt to incorporate semantic information into the limited visual data for category understanding. However, these methods often enrich class-level feature representations with abstract category names, failing to capture the nuanced features essential for effective generalization. To address this issue, we propose a novel framework for FSL, which incorporates both the abstract class semantics and the concrete class entities extracted from Large Language Models (LLMs), to enhance the representation of the class prototypes. Specifically, our framework composes a Semantic-guided Visual Pattern Extraction (SVPE) module and a Prototype-Calibration (PC) module, where the SVPE meticulously extracts semantic-aware visual patterns across diverse scales, while the PC module seamlessly integrates these patterns to refine the visual prototype, enhancing its representativeness. Extensive experiments on four few-shot classification benchmarks and the BSCD-FSL cross-domain benchmarks showcase remarkable advancements over the current state-of-the-art methods. Notably, for the challenging one-shot setting, our approach, utilizing the ResNet-12 backbone, achieves an impressive average improvement of 1.95\% over the second-best competitor. 
\end{abstract}
\section{Introduction} \label{sec:intro}

Deep learning has revolutionized numerous fields, notably computer vision \cite{he2016deep} and natural language processing \cite{achiam2023gpt}. These learning systems are inherently data-intensive, demanding substantial amounts of labeled data for model training. In some practical applications, the acquisition and annotation of data can be prohibitively expensive or unfeasible. Consequently, Few-Shot Learning (\textbf{FSL}) \cite{metaAdaM,zhang2024simple} has garnered significant attention as a promising solution, enabling learning from a scarce quantity of data, thereby mitigating the challenges associated with data scarcity.

Most FSL methods face significant challenges due to the scarcity of samples, especially when only one sample is available for each class \cite{alfa}.
To mitigate this issue, several approaches \cite{yang2022sega,zhang2024simple,sp-clip} have introduced semantic information to assist in constructing class understanding. However, semantic-incorporating methods enrich class representations with high-level information but often overlook the local structure of the visual environment.
Many discriminative patterns in low-level descriptions are crucial for classifying samples, especially when only a few samples are available for each class.

\begin{figure}[!t]
    \centering
    \includegraphics[width=0.98\linewidth]{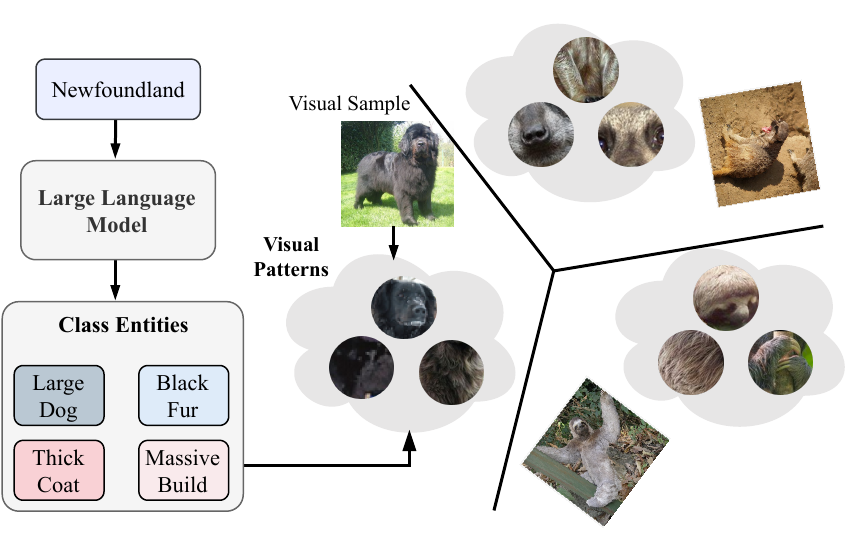}
    \caption{The key idea is to learn a classifier from data using a visual sample per category, extended with additional visual patterns derived from semantic entities generated by LLMs.}
    \label{fig: introduction}
\end{figure}

Humans have demonstrated the ability to recognize novel categories with minimal examples by combining abstract attributes and detailed category descriptions with visual signals and mentally aligning images with this abstract information. If machine vision systems could do such information abstraction and alignment, then the novel categories may be quickly learned with only a few samples. Unfortunately, directly extracting low-level descriptions of categories from visual samples can be challenging, as images often fail to comprehensively capture the full range of attributes and characteristics that define a class. The limitations of visual representation, such as variations in lighting, angles, and occlusion, can hinder the models' capability to comprehensively describe a category based solely on the images at hand. 

Considering the success of nature language processing (NLP) that the related semantic descriptions or attributes could be generated with the power of Large-Language Models (LLMs), we propose a novel paradigm that learns semantic-aware visual patterns from the semantic descriptions to enrich the class representations for FSL, as shown in \cref{fig: introduction}. However, such a paradigm still suffers from challenges during implementation. Though producing some descriptions from the category names, most LLMs often encounter the hallucination problem that the generative descriptions fail to maintain a relevant correspondence with the categories. Further, the misalignment between the visual samples and the low-level descriptions makes it challenging to classify samples.

In this paper, we introduce a novel approach named \textbf{ECER-FSL}, designed to utilize concrete class entities generated by LLMs to enhance FSL. To address the hallucination issue in generating semantic entities for each class, we employ a filtering strategy that evaluates the similarity between the generated semantic descriptions and the corresponding class names, discarding entities with low similarity. Furthermore, to address the critical issue of misalignment, we propose a Progressive Visual-Semantic Aggregation (PVSA) framework that leverages both class names and class-related entities to generate semantic-aware visual patterns, progressively enriching the visual prototypes. The PVSA framework effectively integrates multiple stages of visual features with semantic information via the Semantic-guided Visual Pattern Extraction (SVPE) module, thereby comprehensively uncovering the potential visual characteristics of categories.
By incorporating semantic-aware visual patterns into the visual prototypes through a well-designed Prototype Calibration (PC) module, discriminative classifiers are obtained under limited visual samples. 

Overall, our contributions are as follows:
\begin{itemize}
    \item We introduce a novel FSL paradigm that learns semantic-aware visual patterns from the semantic entities derived from the large-language models to enrich the visual prototype for each category. 

    \item We propose a Progressive Visual-Semantic Aggregation (PVSA) framework that gradually captures the semantic-aware visual patterns at different network blocks guided by both class names and semantic entities. 

    \item Extensive experiments on both FSL and cross-domain FSL tasks have demonstrated that our proposed method establishes a new benchmark for state-of-the-art performance, outperforming the second-best competitor by a substantial margin, specifically under the 1-shot scenario. 
\end{itemize}

\section{Related Works} \label{sec:related}
\subsection{Uni-Modal Few-Shot Learning}
The uni-modal FSL approaches only use the visual support data to predict the query samples. The existing uni-modal FSL approaches could be roughly grouped into three categories, i.e., metric-based, optimization-based, and hallucination-based approaches. Metric-based approaches \cite{snell2017prototypical,chen2019closer} aim to learn a good metric space to evaluate the affinity similarities between the support-query pairs, where the query samples from the novel classes can be nicely categorized via the nearest neighbor classifier. To capture rich statistics, several works \cite{zhang2019variational,DeepBDCCVPR2022} further exploit the second moment to enrich the image representations and perform the similarity metric with different metrics \cite{liu2023cycle, hu2024understanding}. Optimization-based approaches \cite{finn2017model, yu2022multi, sun2024meta} aim to learn how to train model parameters to produce good results on new tasks with a few optimization steps, or even a single optimization step. Hallucination-based approaches address FSL via augmenting the support data at either the image level \cite{zhang2018metagan} or the feature level \cite{lazarou2022tensor,bar2024frozen}. 

\subsection{Multi-Modal Few-Shot Learning} 
Multi-Modal FSL approaches \cite{xing2019adaptive,yang2022sega,ji2022information} learn novel categories from multiple modalities when support visual samples are scarce. These multi-modal methods can complement uni-modal FSL approaches. AM3 \cite{xing2019adaptive} enhances ProtoNet \cite{snell2017prototypical} by integrating class-level semantic and visual prototypes. SEGA \cite{yang2022sega} uses label embeddings to focus on semantic-related visual features. TRAML \cite{TRAML} improves visual feature embeddings by aligning them with semantic similarities. SemFew \cite{zhang2024simple} proposes a simple two-layer network to transform semantic and visual features into robust category prototypes with rich discriminative features. Some studies use semantic features for data augmentation. ProtoComNet \cite{ProtoComNet} generates additional features using a Gaussian distribution based on hand-crafted attribute features. TriNet \cite{TriNet} uses an encoder-decoder framework to align and augment visual features with semantic information. SIFT \cite{sift} uses class-specific semantic embeddings to generate high-quality features via an encoding-transformation-decoding process, enabling models to transfer features from base to novel categories.

Our approach also aligns with multi-modal methods, yet distinguishes itself by integrating expert knowledge from LLMs to enhance visual concept extraction across various stages of the visual encoder.

\section{Methodology} \label{sec:method}
Following the existing FSL literature \cite{Chen_2021_ICCV}, our method encompasses a two-stage process: a pre-training phase that involves learning generalizable representations with the base set, followed by an episode-based fine-tuning stage that the model is further fine-tuned with various FSL tasks generated from the base dataset.

\begin{figure*}[!t]
    \centering
    \includegraphics[width=0.9\linewidth]{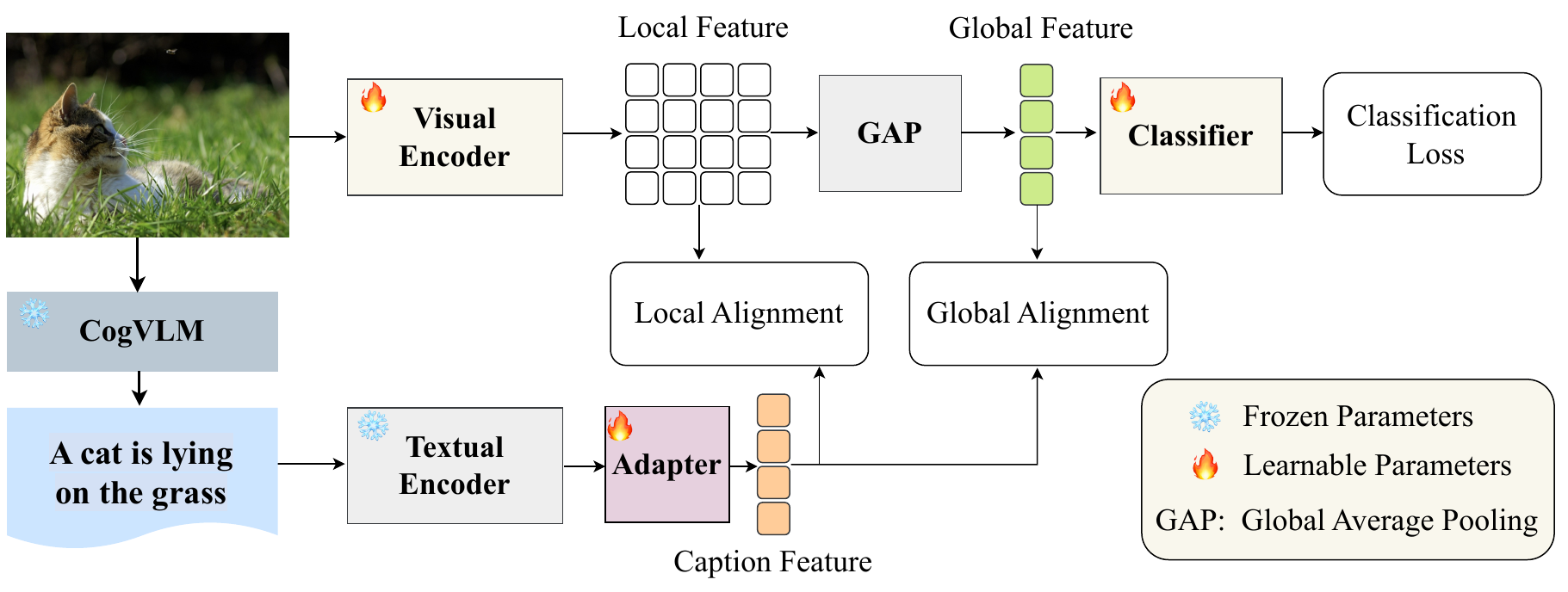}
    \vspace{-2mm}
    \caption{\textbf{Multi-Modality Pre-training Stage}. Our pre-training paradigm trains a visual encoder that captures more semantic-rich information under the guidance of the textual captions derived from an off-the-shelf model, which harnesses the power of natural language processing to enrich the visual representation, fostering a deeper understanding of the images and their underlying semantics. }
\label{fig:pre-training}
\end{figure*}

\subsection{Multi-Modality Pre-training}

As visual feature embedding plays a pivotal role in FSL \cite{tian2020rethinking}, most existing approaches \cite{Chen_2021_ICCV} leverage the available base data to train the visual feature encoder with a proxy classification task. However, such a training paradigm overemphasizes the extraction of classification features related to base classes, which tends to overlook the inherent characteristics of the samples themselves when extracting samples from novel classes. Prior work has attempted to mitigate this limitation by implicitly introducing feature constraints \cite{yang2022few} or instance enhancement \cite{pfsl}. 

Different from the pre-training paradigm in the existing FSL literature that trains the visual backbone in a uni-modality way, we introduce a multi-modality pre-training strategy that leverages both visual and textual data to enhance the representation learning capabilities of the visual backbone. As depicted in \cref{fig:pre-training}, our pre-training paradigm ingeniously harnesses semantic captions generated by the existing caption model CogVLM \cite{wang2023cogvlm} from the input visual samples, where the captions serve as auxiliary information to guide and enhance the visual feature backbone's ability to extract semantic-rich feature representations. Specifically, we apply the off-the-shelf text encoder \cite{clip} to extract the feature embedding and introduce an adapter to handle the compatibility between the visual and text embeddings. To integrate textual information into the pre-training process, we introduce the multi-modality contrastive loss that operates at both local and global levels. The local level focuses on aligning specific regions of the image with corresponding segments of the text, while the global level ensures the overall semantic alignment between the entire image and the text description.

Given an image $I$, the visual model extracts the feature map $\mathbf{F}_i \in \mathbb{R}^{HW \times C}$ and global representation $\mathbf{v}_i \in \mathbb{R}^{1 \times C}$. The caption representation $\mathbf{t}$ of the image $I$ is obtained with the off-the-shelf text encoder followed by an adapter module. The image-text contrastive loss is defined as:
\begin{align} \label{eq: image-text}
\mathcal{L}_{IT} = -&\sum_{\mathbf{I}_i, \mathbf{t}_i \in \mathcal{B}}\log \frac{\exp (\operatorname{sim}(\mathbf{I}_i, \mathbf{t}_i) / \tau)}{\sum_{\mathbf{t}_j \in \mathcal{B}} \mathbb{I}_{j \neq i} \exp(\operatorname{sim}(\mathbf{I}_i, \mathbf{t}_j) / \tau)} \\ \nonumber 
& + \log \frac{\exp (\operatorname{sim}(\mathbf{t}_i, \mathbf{I}_i) / \tau)}{\sum_{\mathbf{I}_j \in \mathcal{B}} \mathbb{I}_{j \neq i} \exp(\operatorname{sim}(\mathbf{t}_i, \mathbf{I}_j) / \tau)},
\end{align}
where $\mathbb{I}$ is the indicator function, $\tau$ represents the temperature parameter, and $\mathcal{B}$ signifies the batch size.

For the local-level visual-semantic alignment, the similarity between the visual and semantic modality is obtained with: 
\begin{align}
    \operatorname{sim}(\mathbf{I}_i, \mathbf{t}_i) = \frac{1}{HW}\sum_{k=1}^{HW} \cos(\mathbf{f}_{i}^{k}, \mathbf{t}),
\end{align}
where $\mathbf{f}_{i}^{k}$ is the $k$-th local feature representation of feature map $\mathbf{F}_i$. For the global-level visual-semantic alignment, the visual-semantic similarity is:
\begin{align}
    \operatorname{sim}(\mathbf{I}_i, \mathbf{t}_i) = \operatorname{cos}(\mathbf{v}_i, \mathbf{t}_i),
\end{align}
where $\cos$ is the cosine similarity. Then the overall loss in a batch size during the pre-training stage is:
\begin{equation}
\mathcal{L}_{\text{pre-trained}} = \mathcal{L}_{CE} + \lambda \mathcal{L}_{IT}^{global} + \eta \mathcal{L}_{IT}^{local},
\end{equation}
where $\mathcal{L}_{CE}$ denotes the cross-entropy loss, $\lambda$ and $\eta$ are hyper-parameters that balance losses. By leveraging this semantic guidance, our approach fosters a deeper understanding of the visual content and strengthens the alignment between visual and semantic modalities, facilitating the model to learn more meaningful and contextually relevant features that better capture the characteristics of the image.

\begin{figure*}[!ht]
    \centering
    \includegraphics[width=\linewidth]{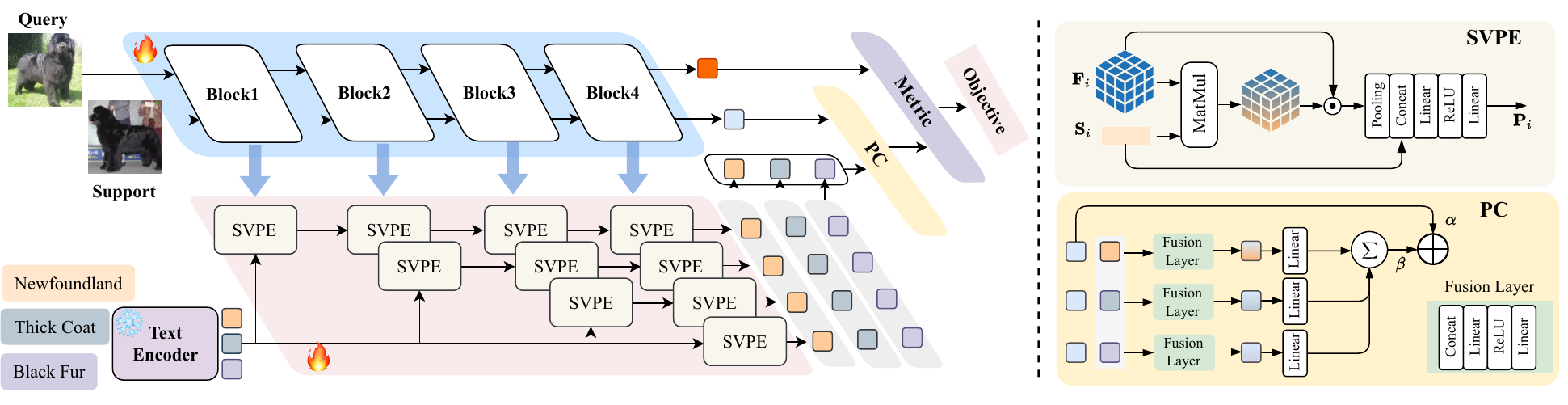}
    \caption{The framework of our progressive visual-semantic aggregation (PVSA) leverages both the class name (``Newfoundland") and class-related entities (``Thick Coat", ``Black Fur") to enrich the visual prototypes. PVSA  consists of a semantic-guided visual pattern extraction (SVPE) module that extracts visual patterns that are related to the class semantics and a prototype calibration (PC) module that enriches the visual prototype by incorporating the semantic-aware visual features. }
    \vspace{-2mm}
    \label{fig:framework}
\end{figure*}

\begin{figure}[!ht]
    \centering
    \includegraphics[width=\linewidth]{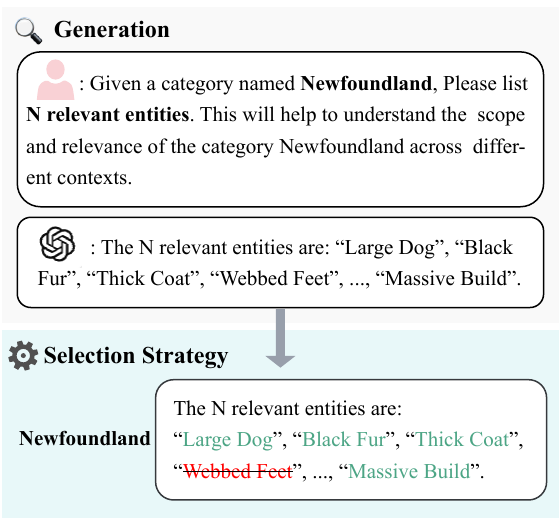}
    \caption{The entities produced with LLMs prompt and filtered with the proposed selection strategy.}
    \vspace{-5mm}
    \label{fig:entity-generation}
\end{figure}

\subsection{Entities-Assisted Episode-based Fine-Tuning}
Given an FSL episode task, which comprises query samples $\mathcal{Q}$ and a support set $\mathcal{S}$ with a limited number of visual samples and corresponding class names for each class, the fine-tuning stage aims to learn discriminative feature prototypes for each class and effective feature representations for the query samples. To extract a representative feature prototype for each support class, we incorporate some prior semantic information related to the key attributes of the categories into the fine-tuning phase. Unlike the abstract conceptualization process that human beings naturally undertake when interpreting visual samples, our approach leverages the derivative and reasoning capabilities of large language models (LLMs) to envision the class entities from the class names, as shown in \cref{fig:entity-generation}.

\textbf{Entity Generation via LLMs.} As depicted in \cref{fig:entity-generation}, the class-specific concepts or entities are first generated by off-the-shelf LLMs and then filtered the irrelevant entities with a selection strategy. Given a class name, \textit{e.g.}, Newfoundland, we elaborate the specific prompts for LLMs to generate $N$ class-related entities, i.e., $E = \{e_n\}_{n=1}^{N}$. However, the generated entities may exhibit hallucinations, leading to the production of irrelevant or unreliable entities. To overcome this limitation, we introduce a selection strategy that assesses the textual similarity between the class name and the generated entities, enabling the identification and selection of only the most relevant entities for inclusion. Given a category name \( t \), we modify it using the prompt ``a photo of a [$t$]" and then calculate the similarity between each entity \( e_i \) and the category name \( t \) using the text encoder \( \theta_{T} \), with the following equation:
\begin{equation} \label{eq: method-select}
    \text{sim}(e_{n}, t) = \operatorname{cos}(\theta_{T}(e_{n}), ~\theta_{T}(t)),
\end{equation}
where $\operatorname{cos}$ represents the cosine similarity metric. Then, we rank the entities based on their similarity scores and select the top-$K$ entities as the class-specific entities.

\subsection{Progressive Visual-Semantic Aggregation}

As illustrated in \cref{fig:framework}, we introduce a progressive visual-semantic aggregation (PVSA) framework that leverages both the class name and class-related entities generated by the LLMs to enrich the visual prototypes.
This focuses on concentrating class-specific visual features at each stage of information extraction, allowing for more refined visual patterns through semantic guidance.
Specifically, PVSA mainly consists of a semantic-guided visual pattern extraction module and a prototype calibration module. 

\subsubsection{Semantic-guided Visual Pattern Extraction (SVPE).}
SVPE is the key module to extract visual patterns that are related to the class names or class-wise entities. Given the feature map $\mathbf{F}_{i} \in \mathbb{R}^{HW \times D}$ from $i$-th block, the semantic-aware visual feature $\mathbf{P}_i$ is obtained with a $\operatorname{SVPE}$ module, which is formulated with: 
\begin{equation}
    \begin{aligned}
        \mathbf{P}_i = \operatorname{SVPE}(\mathbf{S}_i, \mathbf{F}_i).
    \end{aligned}
\end{equation}

For the first block, $\mathbf{S}_i$ is the feature embedding from the semantic feature set that consists of the semantic entities $\mathbf{T}_k (k \leq K)$ and class names $\mathbf{T}_{cls}$. For the other block, $\mathbf{S}_i$ is the feature embedding from the joint set that consists of both semantic features and the semantic-aware visual feature $\mathbf{P}_{i-1}$ from the previous block. Specifically, the SVPE module is formulated as:
\begin{equation}
\begin{aligned} \label{eq: semattn}
    \mathbf{F}_{i}' &= \operatorname{Pooling}(\operatorname{MatMul}(\mathbf{S}_i, \mathbf{F}_{i}) \cdot \mathbf{F}_{i}), \\ 
    \mathbf{P}_{i} &= \operatorname{FusionLayer}(\mathbf{F}_{i}', ~\mathbf{S}_i),
\end{aligned}
\end{equation}
where $\operatorname{FusionLayer}$ integrates input pairs $\mathbf{F}_{i}', ~\mathbf{S}_i$ by first concatenating and then processing them through a two-layer network with LeakyReLU activation.

Once the semantic-aware visual features about the specific semantic entity or class name from the final block of the visual backbone have been obtained, they are then concatenated with the visual support prototypes for further refinement. For $k$-th semantic entity, we obtain:  
\begin{equation} \label{eq: prototype fusion}
    \mathbf{P}_{entity}^k = \operatorname{FusionLayer}(\{\mathbf{P}_{entity\mbox{-}{k}}^n\}_{n=1}^{4}),
\end{equation}
where $\operatorname{FusionLayer}$ is a concatenation operation followed by a two-layer network. Thus, we obtain the corresponding class-level semantic prototype as $\mathbf{P}_{cls}$ and the entity-level semantic prototypes as $\mathbf{P}_{entity} = \{\mathbf{P}_{entity}^{k} \mid k=1,\dots,K\}$, $K$ is the entity number.

\subsubsection{Prototype Calibration.} The prototype calibration (PC) module incorporates diverse semantic-aware visual features into the feature embedding of the support samples, thereby generating a refined, calibrated visual prototype that captures a more comprehensive and semantically enriched representation. The process is formulated as: 

\begin{equation} \label{eq:late-fusion}
\begin{aligned}
    &\mathbf{C}_{cls} = \operatorname{FusionLayer}(\mathbf{f}^s, \mathbf{P}_{cls}), \\
    &\mathbf{C}_{entity}^k = \operatorname{FusionLayer}(\mathbf{f}^s, \mathbf{P}_{entity}^k), \\
    &\mathbf{C} = \alpha \cdot \mathbf{f}^s + \beta \cdot (\operatorname{FC}(\mathbf{C}_{cls}) + \sum_{k=1}^{K} \operatorname{FC}(\mathbf{C}_{entity}^{k})),
\end{aligned}
\end{equation}
where $\mathbf{f}^{s} = \operatorname{AveragePooling}(\mathbf{F}^{N})$ represents the average-pooled feature from the final block, $\mathbf{P}_{cls}$ and $\mathbf{P}_{entity}$ respectively denote the class fusion prototype and entities fusion prototype obtained with \cref{eq: prototype fusion}. $\operatorname{FC}$ denotes a fully connected layer for feature dimensionality mapping. $\alpha$ and $\beta$ are two hyper-parameters.

To this end, the model culminates in the generation of a comprehensive representative prototype, denoted as $\mathbf{C}$, which seamlessly integrates both semantic and visual features. The multi-layered fusion strategy, where semantic and visual signals are harmoniously combined at each stage, empowers our model to leverage the full potential of both information streams. 

\begin{table*}[!h]
    \centering
    \resizebox{0.99\linewidth}{!}{
    \setlength{\tabcolsep}{2.4mm}
    \begin{tabular}{c|l l c c c cc}
    \toprule
        & Method & Backbone & MiniImageNet & TieredImageNet & CIFAR-FS & FC100 & Average\\
        \midrule
        \multirow{7}{*}{\rotatebox{90}{\textbf{Visual}}}
        & SUN \cite{sun} & ViT-S & 67.80 ± 0.45 & 72.99 ± 0.50 & 78.37 ± 0.46 & - & -\\
        &FewTURE \cite{fewture} & Swin-T & 72.40 ± 0.78 & 76.32 ± 0.87 &77.76 ± 0.81  & 47.68 ± 0.78 & 68.54\\
        & FGFL \cite{FGFL} & ResNet-12 & 69.14 ± 0.80 & 73.21 ± 0.88 & - & - & -\\
        & CPEA \cite{CPEA} & ViT-S & 71.97 ± 0.65 & 76.93 ± 0.70 &77.82 ± 0.66 & 47.24 ± 0.58 & 68.49\\
        &Meta-AdaM \cite{metaAdaM} & ResNet-12 & 59.89 ± 0.49  & 65.31 ± 0.48 & -  & 41.12 ± 0.49 & -\\
        & ALFA \cite{alfa} & ResNet-12 & 66.61 ± 0.28 & 70.29 ± 0.40 & 76.32 ± 0.43 & 44.54 ± 0.50 & 64.44\\
        & LastShot \cite{last-shot} & ResNet-12 & 67.35 ± 0.20 & 72.43 ± 0.23 & 76.76 ± 0.21 & 44.08 ± 0.18 & 65.16\\  
        \midrule
        \multirow{6}{*}{\rotatebox{90}{\textbf{Semantic}}}
        &SVAE-Proto \cite{svae-proto} & ResNet-12 & 74.84 ± 0.23 & 76.98 ± 0.65 & -&- &-\\
        &SP-CLIP \cite{sp-clip} & Visformer-T & 72.31 ± 0.40 & 78.03 ± 0.46 & 82.18 ± 0.40 & 48.53 ± 0.38 & 70.26\\ 
        & SIFT \cite{sift} & WRN-28-10 & 77.31 ± 0.67 & 77.86 ± 0.77 & 81.35 ± 0.75 & - & -\\
        & Sem-Few \cite{zhang2024simple} & ResNet-12 & 77.63 ± 0.63 & 78.96 ± 0.80  & 83.65 ± 0.70 & 54.36 ± 0.71 & 73.65\\
        \cmidrule(lr){2-8}        
        & ECER-FSL (Ours) & ResNet-12 & \underline{80.34 ± 0.21}  & \underline{80.79 ± 0.57} & \underline{85.13 ± 0.61} & \underline{56.12 ± 0.41} & \underline{75.60}\\
        & ECER-FSL (Ours) & Visformer-T & \textbf{81.14 ± 0.15}  & \textbf{81.81 ± 0.51} & \textbf{86.01 ± 0.35}  & \textbf{57.34 ± 0.31}  & \textbf{76.58}\\
        \bottomrule
    \end{tabular}}
    \caption{Results (\%) on four datasets on \textbf{5-way 1-shot} tasks. The ± shows 95\% confidence intervals. ``Visual" and ``Semantic" indicate the visual-only-based and semantic-incorporated-based methods. The results for the competitors are directly from the published literature. The best and second-best results are shown in \textbf{bold} and \underline{underline}, respectively.}
    \label{tab:maincomparison}
\end{table*}

\subsubsection{Objective Function.} The probability of the query sample $q$ to the $i$-th class can be obtained with:  
\begin{equation} \label{eq:cosine}
p_i=\frac{\exp \left(\operatorname{cos} \left(f\left(q_i\right), \mathbf{C}_i\right)\right / \tau)}{\sum_{j=1}^N \exp \left(\operatorname{cos} \left(f\left(q_i\right), \mathbf{C}_j\right)\right / \tau)},
\end{equation}
where $f(q_i)$ is the feature embedding of query sample $q$, $\operatorname{cos}$ denotes the cosine similarity and $\tau$ is the temperature ratio. The objective function is defined with a cross-entropy loss:
\begin{equation} \label{eq: loss}
\mathcal{L} = \sum_{i=1}^M \operatorname{CrossEntropy}(p_{i}, y_i),
\end{equation}
where $y_i$ donates the corresponding ground-truth label of the query image $q_i$, and $M$ is the number of query samples in each episode.

\section{Experiments} \label{sec:exp}
\definecolor{tabcolor}{RGB}{200, 200, 200}
\def\tabwidth{.31}

\subsection{Experimental Details}

\textbf{Datasets.}
We evaluate the proposed method across two primary tasks: the traditional FSL and the cross-domain FSL (CD-FSL). The traditional FSL is evaluated on four datasets, namely \textbf{MiniImageNet} \cite{vinyals2016matching}, \textbf{TieredImageNet} \cite{ren2018meta}, \textbf{CIFAR-FS} \cite{lee2019meta}, and \textbf{FC100} \cite{oreshkin2018tadam}.
Following \cite{BSCD-FSL}, we evaluate the CD-FSL on BSCD-FSL benchmark, which involves training on MiniImageNet and testing on four unrelated datasets: \textbf{ChestX} \cite{chestX}, \textbf{ISIC} \cite{ISIC}, \textbf{EuroSAT} \cite{EuroSAT}, and \textbf{CropDisease} \cite{CropDisease}.

\textbf{Training Details.} We employ two vision backbones for comparison: ResNet12 \cite{he2016deep} and Visformer-T \cite{chen2021visformer}. For text encoding, we use ViT-B/32 CLIP \cite{clip} and we utilize GPT-4-o \cite{ChatGPT} to generate related entities. The pre-training stage is set to 200 epochs for all datasets, while the meta-training stage is set to 50 epochs. 
The $\alpha$ and $\beta$ in \cref{eq:late-fusion} are consistently assigned values of 0.2 and 0.8, respectively, across all datasets. During the pre-training phase, we set the batch size to 128, leveraging the Adam optimizer \cite{kingma2014adam} with a learning rate of 1e-4 for optimization of the model parameters.

\textbf{Evaluation Protocol.} For the evaluation, we uniformly sampled 600 classification tasks from a novel set that comprises classes that are disjoint from those in the base set. 
In each task, there are 15 query samples for each class. The mean and 95\% confidence interval of the accuracy are reported.

\subsection{Performance Comparison}
\textbf{Few-Shot Classification. }
Table~\ref{tab:maincomparison} provides a comprehensive comparison of the performance of our method against both visual-only-based and semantic-incorporated-based competitors under the 5-way 1-shot task across four benchmarks. From the results, we observe that our method achieves the best with both CNN and Transformer backbones on four datasets, outperforming the second-best competitor Sem-Few \cite{zhang2024simple} by significant margins. We speculate that the superior performance is due to our method capturing semantic-rich feature representations for each support class. Besides, we observe that the methods incorporating semantic information exhibit significantly superior performance compared to those solely reliant on visual cues, suggesting that prior semantic knowledge can serve as valuable hints, enhancing the overall effectiveness and accuracy of the tasks at hand. Moreover, the results obtained with the Visformer-T backbone perform better than those obtained with the ResNet backbone.

\textbf{Cross-Domain Classification.}
\cref{tab:bscd} shows the comparison results of our method and six competitors on the BSCD-FSL benchmark. Our approach demonstrates a substantial advantage over the other competitors, achieving a notable improvement ranging from 1.39\% to 3.71\% over the second-best performing method across four diverse datasets. This underscores the significant value of incorporating rich semantic information in cross-domain scenarios.

\begin{table}[t]
    \centering
    \resizebox{\linewidth}{!}{
    \begin{tabular}{lcccc}
    \toprule
       \textbf{Method} & \textbf{\small{ChestX}} & \textbf{\small{ISIC}} & \textbf{\small{EuroSAT}} & \textbf{\small{CropDisease}} \\
        \midrule
        GNN \cite{gnn} & 22.00 & 32.02 & 63.69 & 64.48 \\
        ATA \cite{ATA} & 22.10 & 33.21 & 61.35 & 67.47 \\ 
        AFA \cite{AFA} & 22.92 & 33.21 & 63.12 & 67.61 \\
        StyleAdv \cite{fu2023styleadv} & 22.64 & 33.96 & \underline{70.94} & 74.13\\
        LDP-net \cite{ldp-net} & \underline{23.01} & 33.97 & 65.11 & 69.64 \\
        Dara \cite{dara2023} & 22.92 & \underline{36.42} & 67.42 & \underline{80.74} \\
    \midrule
        ECER-FSL (Ours) & \textbf{25.12} & \textbf{40.13} & \textbf{74.13} & \textbf{82.13} \\
    \midrule
        $\triangle$ & 2.11 & 3.71 &  3.19 & 1.39\\
    \bottomrule
    \end{tabular}
    }
    \caption{Average results (\%) on BSCD-FSL benchmarks. $\triangle$ denotes ECER-FSL's gain over the second-best competitors.}
    \label{tab:bscd}
\end{table}

\subsection{Ablation Study}
To comprehensively evaluate the effectiveness of our method, we conduct experiments on both MiniImageNet and TieredImageNet datasets with ResNet12 as the backbone.


\begin{table}[!t]
    \centering
    \resizebox{\linewidth}{!}{
    \begin{tabular}{cccc|cc}
    \toprule
    \textbf{{$\mathcal{L}_{CE}$} }& \textbf{{$\mathcal{L}_{global}$}} & \textbf{{$\mathcal{L}_{local}$}} & {\textbf{\textit{Adapter}}} & \textbf{MiniImageNet} & \textbf{TieredImageNet}  \\    
    \midrule    
    \cmark & \xmark & \xmark & \xmark & 78.30 ± 0.38 & 79.10 ± 0.47 \\ 
    \cmark & \cmark & \xmark & \xmark & 79.21 ± 0.29 & 79.35 ± 0.62 \\    
    \cmark & \cmark & \cmark & \xmark & 79.78 ± 0.31 & 80.32 ± 0.27 \\ 
    \cmark & \cmark & \cmark & \cmark & 80.34 ± 0.21 & 80.79 ± 0.57 \\ 
    \bottomrule
    \end{tabular}
    }
    \caption{Ablation results (\%) of during pre-training stage.}
    \label{tab:ab-pre-training}
\end{table}

\textbf{Ablation Study during Pre-training.}
\cref{tab:ab-pre-training} shows the ablation results during the pre-training stage. From the results, we observe that our multi-modality pre-training method performs a substantial improvement over the baseline model trained solely with CE loss. Specifically, combined $\mathcal{L}_{CE}$ with $\mathcal{L}_{global}$ improves the accuracy by an average of 0.91\% and 0.25\% on MiniImageNet and TieredImageNet, respectively.
Meanwhile, the addition of $\mathcal{L}_{local}$ further enhances the performance, demonstrating that semantic alignment of local information helps feature representation learning. Notably, we observe an even greater enhancement in performance when employing the adapter to map textual features to visual representations, as opposed to the reverse direction. 

\textbf{Ablation Study during Fine-tuning.}
\cref{tab: ab-design} demonstrates that each module in our proposed PVSA framework during the fine-tuning stage. By incorporating the SVPE during the fine-tuning phase, we achieve substantial performance gains, with notable improvements of 12.91\% and 10.78\% on minImageNet and TieredImageNet, respectively. This suggests that progressively fusing semantic features into visual signals facilitates a better understanding of related class concepts. Moreover, the inclusion of the PC module significantly amplifies performance. Besides, the integration of entity concepts within this framework yields an additional leap in performance.

\begin{table}[!t]
    \centering
\resizebox{\linewidth}{!}{
    \begin{tabular}{lcc|cc}
    \toprule
    {\textbf{SVPE}} & {\textbf{PC}} & {\textbf{Entity}} & \textbf{MiniImageNet} & \textbf{TieredImageNet} \\
    \midrule
    \xmark & \xmark & \xmark & 65.39 ± 0.29 & 68.32 ± 0.78 \\
    \cmark & \xmark & \xmark & 78.30 ± 0.38 & 79.10 ± 0.47 \\ 
    \cmark & \cmark & \xmark & 79.34 ± 0.33 & 79.51 ± 0.49 \\
    \cmark & \xmark & \cmark & 79.92 ± 0.27 & 80.21 ± 0.29 \\
    \cmark & \cmark & \cmark & 80.34 ± 0.21 & 80.79 ± 0.57 \\
    \bottomrule
    \end{tabular}
}
    \caption{Ablation results (\%) during fine-tuning stage.}
    \label{tab: ab-design}
\end{table}

\begin{figure}[!t]
    \centering
    \begin{subfigure}[b]{0.491\linewidth}
        \centering
        \includegraphics[width=\linewidth]{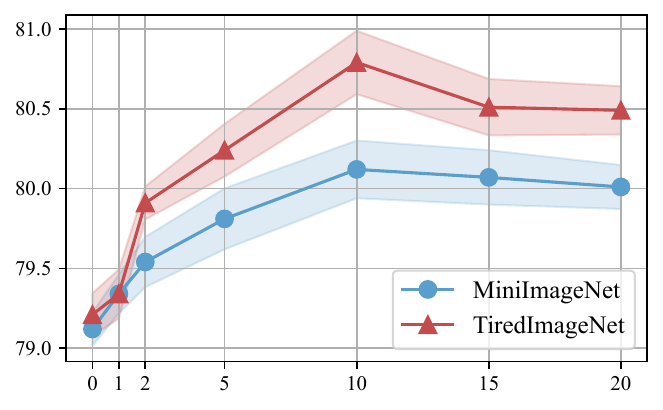}
        \vspace{-5mm}
        \caption{\small{Entity Numbers}}
        \label{fig:ab-entity-a}
    \end{subfigure}
    \hfill
    \begin{subfigure}[b]{0.491\linewidth}
        \centering
        \includegraphics[width=\linewidth]{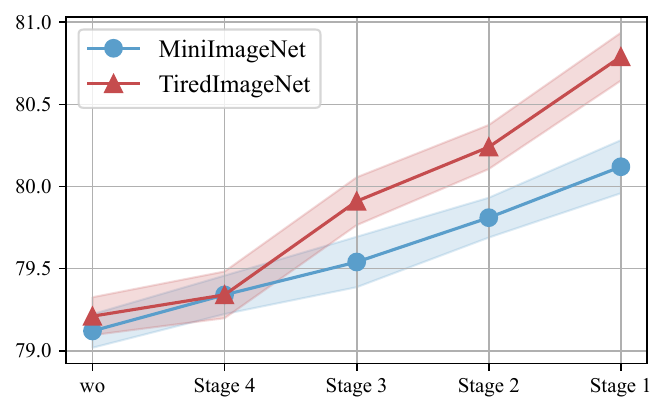}
        \vspace{-5mm}
        \caption{\small{SVPE Stages}}
        \label{fig:ab-entity-b}
    \end{subfigure}
    \vspace{-5mm}
    \caption{Effects (\%) of entity number and SVPE stages.}
    \label{fig:ab-entity}
\end{figure}

\textbf{Impacts of Entity Numbers and SVPE Inserted Stages.} In this experiment, we evaluate how the number of both entities and the SVPE modules inserted in the framework affected the performance of the method for the FSL task. To do so, we varied the number of entities from 0 to 20 and the number of SVPE modules. According to \cref{fig:ab-entity}, we can observe that the performance is enhanced with an increased number of entities and plateaus when the number of entities is above 10, possibly because there is no further additional class information. Thus, the optimal number of entities is therefore set to 10.
Furthermore, as the number of stages where semantic information is infused into the model increases from 1 to 4 blocks, we observe a consistent and incremental improvement in performance. This suggests that incorporating semantic information at multiple stages within the model architecture contributes positively to enhancing its overall capabilities.

\subsection{Further Analysis}
\begin{figure}[!t]
    \centering
    \begin{overpic}[width=\linewidth]{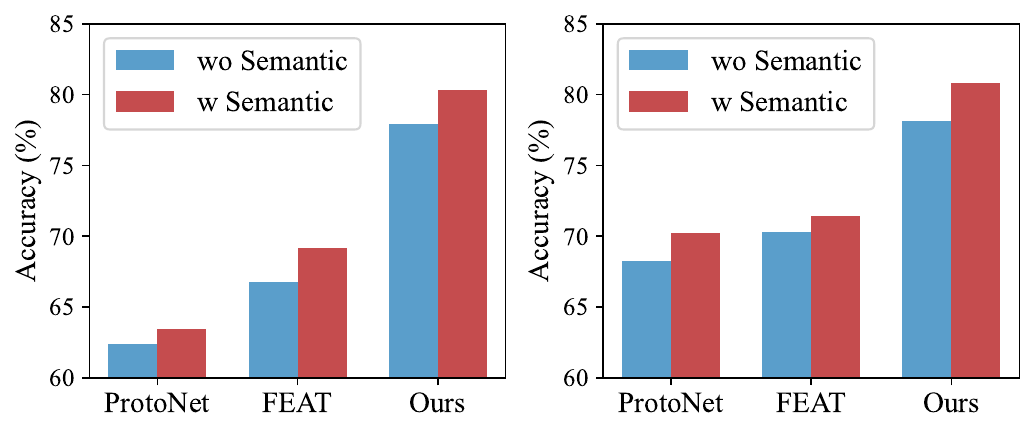}
    \put(14,-3){\small{(a) MiniImageNet}}
    \put(62,-3){\small{(b) TieredImageNet}}
    \end{overpic}
    \vspace{-2mm}
    \caption{Experimental results of combination FSL method with our pre-training model. }
    \label{fig:backbone}
\end{figure}

\textbf{Combination FSL Competitors with Our Pre-training Model.} In this experiment, we evaluate the performance when combining our pre-trained visual backbone with the existing FSL methods ProtoNet \cite{snell2017prototypical} and FEAT \cite{feat}. As depicted in \cref{fig:backbone}, it is evident that the existing methods, when coupled with our pre-training strategy, exhibit significantly superior performance compared to those utilizing a pre-training approach that solely focuses on a visual classification task. This advantage is consistent across both the miniImageNet and TieredImageNet datasets, underscoring the effectiveness of our pre-training strategy.

\begin{figure}[!t]
    \centering
    \begin{overpic}[width=\linewidth]{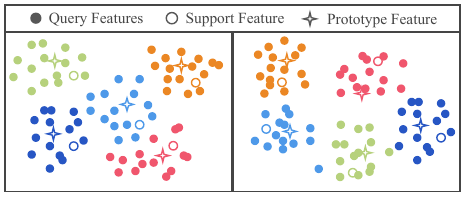}
    \put(12,-3){\small{(a) MiniImageNet}}
    \put(58,-3){\small{(b) TieredImageNet}}
    \end{overpic}
    \vspace{-2mm}
    \caption{Visualization of a random 5-way 1-shot task, where 15 query images and 1 support image for each test class from both MiniImageNet and TieredImageNet datasets.}
    \label{fig:t-sne}
\end{figure}

\textbf{t-SNE Visualization.} To further evaluate the impacts of our method in the feature embedding space, we visualize a random 5-way 1-shot test task sampled from both 
MiniImageNet and TieredImageNet datasets in \cref{fig:t-sne}. Each class contains 15 query samples, 1 support sample, and 1 class prototype. The features of query and support samples are derived from the visual backbone, and the class prototype is after the semantic fusion of PSVF and PC. The visualization results indicate that the refined class prototypes can represent class concepts more effectively than those obtained with the original limited visual samples.

\begin{figure}[!t]
    \centering
    \includegraphics[width=\linewidth]{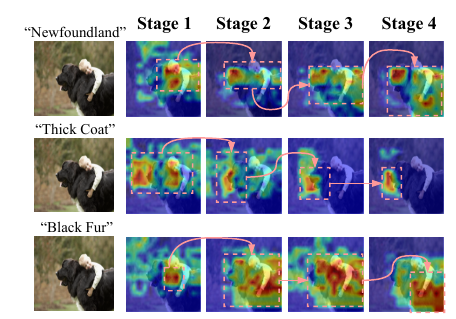}
    \caption{The feature map is computed based on the similarity between feature maps at different stages and the corresponding semantic features. The three rows correspond to ``Newfoundland", ``Thick Coat", and ``Black Fur" respectively.}
    \label{fig: feature-map}
\end{figure}

\textbf{Feature Map Visualization.} \cref{fig: feature-map} shows the feature map. In this experiment, we further evaluate the visualization results at different network blocks under different semantic entities and the class name (``Newfoundland"). The provided visual sample may contain extraneous elements, such as people, grass, and other irrelevant details. From the result, our method effectively extracts information across a hierarchy of features, from low-level to high-level, allowing class-related entities to distill pertinent visual concepts and maintain a sharper focus.
This is evident in distinct regions, with respective feature maps undergoing targeted optimization at different stages.

More experimental results are provided in the Supplementary Materials.

\section{Conclusion} \label{sec:conclusion}
In this paper, we have revisited the pivotal role of semantics in the domain of few-shot learning (FSL). Our proposed model, ECER-FSL, has been designed to harness the full spectrum of semantic information. By leveraging the powerful expert knowledge in LLMs, we develop a strategy to generate and filter the class-related entities for constructing the comprehensive class concept. Additionally, by leveraging the progressive semantic-visual aggregation, ECER-FSL has demonstrated its capability to generate representative class prototypes, which are instrumental in enhancing classification accuracy, particularly in one-shot learning tasks. Experimental results on both the FSL task and cross-domain FSL task have shown the effectiveness of our ECER-FSL. \bibliography{aaai25}

\end{document}